\pdfoutput=1

\documentclass[11pt]{article}

\usepackage[preprint]{acl}
\usepackage{tabularx}
\usepackage{array}
\usepackage{float}
\newcolumntype{C}[1]{>{\centering\arraybackslash}p{#1}}
\usepackage{times}
\usepackage{latexsym}
\usepackage{amsmath}
\usepackage{amssymb}
\usepackage{enumitem}
\usepackage[T1]{fontenc}

\usepackage[utf8]{inputenc}

\usepackage{microtype}

\usepackage{inconsolata}

\usepackage{graphicx}

\title{Profile-LLM: Dynamic Profile Optimization for Realistic Personality Expression in LLMs}

\author{
 \textbf{Shi-Wei Dai\textsuperscript{1}},
 \textbf{Yan-Wei Shie\textsuperscript{1}},
 \textbf{Tsung-Huan Yang\textsuperscript{1}},
 \textbf{Lun-Wei Ku\textsuperscript{1}},
\\
 \textbf{Yung-Hui Li\textsuperscript{2}}
\\
\\
 \textsuperscript{1}Academia Sinica, Taipei, Taiwan
\\
 \textsuperscript{2}AI Research Center, Hon Hai Research Institute, Taipei, Taiwan
\\
 \small{
   \textbf{Correspondence:} \href{mailto:khrisintw@gmail.com}{khrisintw@gmail.com}
 }
}

\begin{document}
\maketitle
\begin{abstract}
Personalized Large Language Models (LLMs) have been shown to be an effective way to create more engaging and enjoyable user-AI interactions. While previous studies have explored using prompts to elicit specific personality traits in LLMs, they have not optimized these prompts to maximize personality expression. To address this limitation, we propose PersonaPulse: Dynamic Profile Optimization for Realistic Personality Expression in LLMs, a framework that leverages LLMs' inherent knowledge of personality traits to iteratively enhance role-play prompts while integrating a situational response benchmark as a scoring tool, ensuring a more realistic and contextually grounded evaluation to guide the optimization process. Quantitative evaluations demonstrate that the prompts generated by PersonaPulse outperform those of prior work, which were designed based on personality descriptions from psychological studies. Additionally, we explore the relationship between model size and personality modeling through extensive experiments. Finally, we find that, for certain personality traits, the extent of personality evocation can be partially controlled by pausing the optimization process. These findings underscore the importance of prompt optimization in shaping personality expression within LLMs, offering valuable insights for future research on adaptive AI interactions.

\end{abstract}

\section{Introduction}

 \begin{figure*}[t] 
     \centering
     \includegraphics[width=0.8\textwidth]{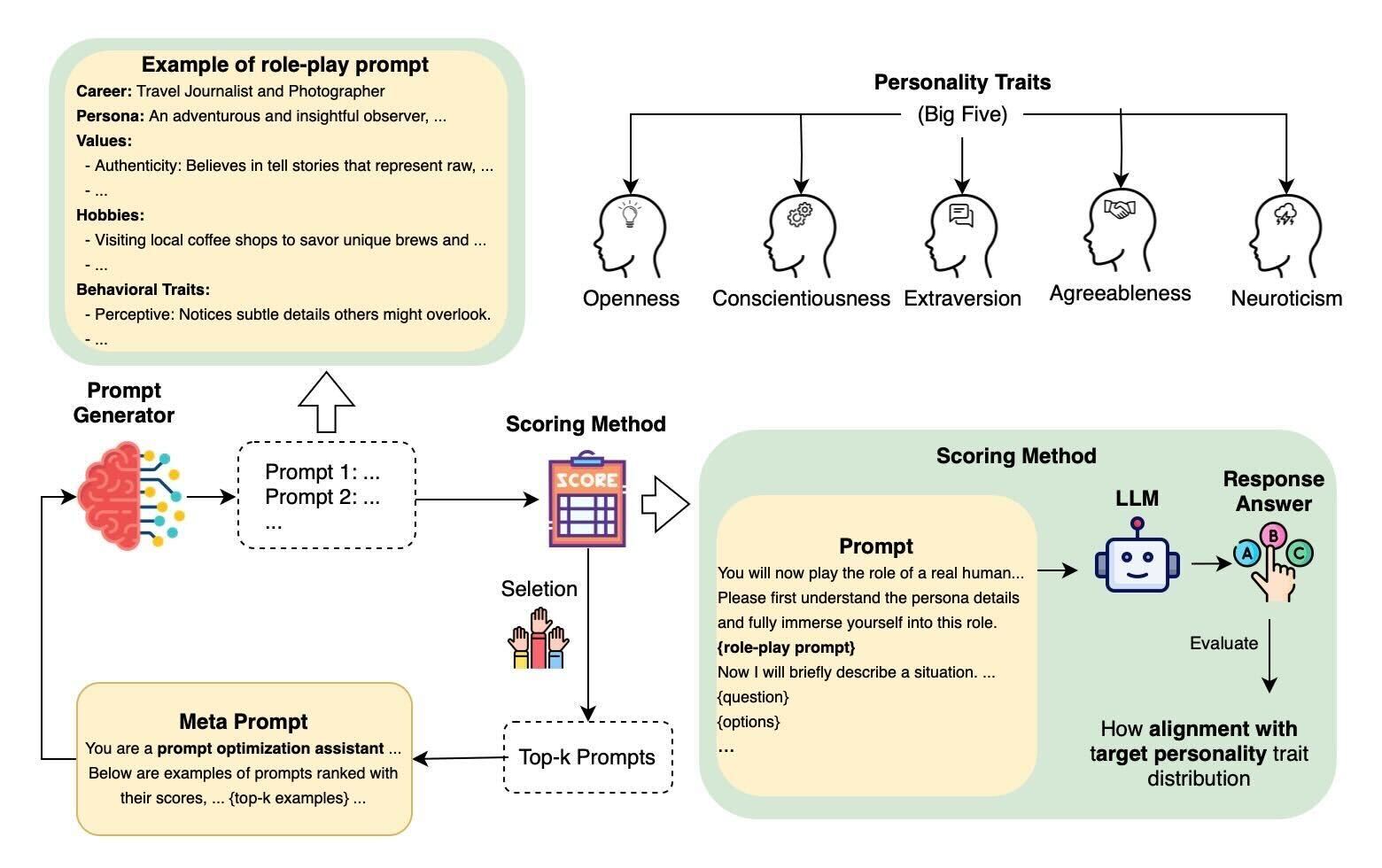} 
     \caption{Profile-LLM framework}
     \label{frame_work}
 \end{figure*}

Personalized Large Language Models (LLMs) can significantly enhance user-AI interactions by adapting their communication style, tone, and responses to align more closely with user expectations and preferences~\citep{salemi-etal-2024-lamp}. Such adaptability is critical in contexts where personality expression strongly influences user engagement. For example, users commonly prefer chatbots exhibiting extraverted traits over neurotic ones and favor conscientious service bots for recommendation tasks~\citep{jiang2023evaluating}. Likewise, tailoring robot personalities to match users' extroverted or introverted traits can notably improve patient engagement and recovery outcomes~\citep{tapus2008user}.

Recent research~\citep{jiang2023evaluating, serapiogarcía2025personalitytraitslargelanguage, lee-etal-2025-llms} has explored methods to measure and validate personality traits in LLMs, confirming their empirical existence. These findings underline the need for effective techniques to evoke personality characteristics in LLMs.

To further refine LLM personalities for diverse applications, recent studies have examined various methods, including parameter modification~\citep{vu2025psychadapter} and prompt engineering~\citep{wen2024selfassessment}. PsychAdapter~\citep{vu2025psychadapter}, for instance, integrates personality scores directly into Transformer architectures, enabling text generation that aligns closely with specific personality traits. Despite its effectiveness, this fine-tuning method's limited transferability and dependence on direct access to model weights reduce its practical applicability. Consequently, prompting has become increasingly popular~\citep{jiang2023evaluating}, especially as it accommodates black-box LLMs available only via inference APIs, such as GPT-4~\citep{openai2023gpt4}. Notably, TRAIT~\citep{lee-etal-2025-llms} employs descriptive prompts based on personality traits from the Big Five Inventory (BFI), while \citet{jiang2023evaluating} utilize self-generated descriptive prompts to elicit specific personality behaviors.

However, current prompting methods~\citep{serapiogarcía2025personalitytraitslargelanguage, jiang-etal-2024-personallm} predominantly rely on abstract, human-crafted or algorithmically generated personality descriptions, limiting their effectiveness in guiding precise personality-specific behaviors in LLMs. Additionally, their static nature inhibits iterative refinement. To address these gaps, we introduce \textbf{Profile-LLM}, a framework that utilizes refined persona profiles to systematically evoke LLM personalities, as depicted in Figure \ref{frame_work}. A persona profile, incorporating detailed characteristics such as career, personality traits, values, hobbies, and behavioral patterns, leverages intuitive and easily comprehensible human concepts of profiling and role-play. This concrete, fine-grained approach enhances controllability and practical adaptability for downstream applications like education, therapy, and entertainment.

To optimize these persona profiles, we iteratively refine them using OPRO~\citep{yang2024large}, with TRAIT~\citep{lee-etal-2025-llms} providing a situational response benchmark to evaluate and guide the optimization process. Extensive experiments demonstrate that larger LLMs more effectively capture and express complex personality traits compared to smaller models, highlighting model size as an influential factor in personality modeling.

The contributions of this paper are three fold.
\begin{enumerate}[left=0pt, itemsep=1pt, parsep=0pt, topsep=0pt]
\item Introducing \textbf{Profile-LLM}, a novel framework to refine and evoking LLM personalities, and demonstrating its superior performance in eliciting personality expressions for mid-sized models compared to existing methods.
\item Investigating the transferability and adaptability of optimized profiles in different LLM models.
\item Exploring methods to modulate the degree of personality expression, thereby offering practical insights for adaptive and personalized AI applications.
\end{enumerate}

\section{Related Work}
\subsection{Personality Modeling in Natural Language}

%

Linguistic theory and psycholinguistic evidence agree that personality systematically manifests in text: word choice, syntactic patterns, and discourse markers all correlate with Big-Five traits \citep{mairesse-walker-2007-personage}.
This observation motivated a line of work on \emph{predicting} personality from language.  Open-vocabulary analyses of Facebook and Twitter showed that lexical cues alone can recover Big-Five scores, sometimes outperforming human judges \citep{Kosinski+2013, schwartz2013personality, youyou2015computer}, and deep models later generalized these findings across languages \citep{liu-etal-2017-language}.
With the advent of large-scale pretraining, researchers began to view LLMs as vast aggregates of human text whose internal representations encode rich personality features.  Early studies demonstrated that role-playing prompts could \emph{evoke} recognizable traits in GPT-like models \citep{jiang2023evaluating, saha-etal-2022-stylistic}.
To measure such behavior automatically, psychometric testbeds such as MPI and TRAIT were proposed; both exhibit validity and reliability comparable to those obtained when testing humans \citep{serapiogarcía2025personalitytraitslargelanguage, lee-etal-2025-llms}.
These tools revealed that LLM personalities are steerable: chain prompting, intent conditioning, and trait-adjective cues can bias responses toward target traits \citep{serapiogarcía2025personalitytraitslargelanguage, sorokovikova-etal-2024-llms}.  Yet steerability alone does not guarantee \emph{controllability}: existing prompts are fixed once written and cannot be continuously tuned for a desired intensity level.

\subsection{Structured Persona-Profile Prompting}
To improve consistency beyond single-sentence cues, later work adopted structured profiles.  
\citet{Huang+2024} conditioned instruction-tuned models on Big-Five vectors, while RoleLLM \citep{wang-etal-2024-rolellm} fine-tuned with multi-field persona sheets, yielding more coherent role play.
Further studies confirmed that LLMs can faithfully simulate assigned profiles across tasks \citep{jiang2023evaluating, sorokovikova-etal-2024-llms}; memory-augmented agents demonstrated realistic long-term interactions \citep{Park2023GenerativeAgents}.
However, these profiles are inherently \emph{static}: once drafted—whether by humans or upstream LLMs—they cannot be adapted on-the-fly to produce finer or subtler expressions of a target trait.

\subsection{Prompt Engineering and Optimization}
Two families of methods attempt to go beyond static prompts.  

\paragraph{Parameter-level tuning.}
Techniques such as soft prompts~\citep{lester2021power}, prefix-tuning~\citep{li-liang-2021-prefix}, LoRA~\citep{hu2021lora}, and RLHF/PPO fine-tuning~\citep{christiano2017deep, ouyang2022training}, as well as PsychAdapter~\citep{vu2025psychadapter}, directly modify model parameters, achieving strong alignment but at the cost of gradient access, substantial compute, and poor cross-model transferability.

\paragraph{Black-box prompt search.}
When weights are frozen, optimization operates purely in text space.
Automatic Prompt Engineering (APE) iteratively generates and ranks full prompts \citep{zhou2023large}; EvoPrompt \citep{guo2024connecting} and Promptbreeder \citep{fernando2024promptbreeder} evolve token sequences via crossover and mutation.  
These approaches perform well on generic tasks, yet they struggle with structured persona profiles: APE rewrites the entire prompt wholesale, while token-level mutations often disrupt semantics, and none retains a full optimization trajectory—making it hard to pick an exact \emph{stop point} for a chosen trait strength.

\paragraph{OPRO for profile refinement.}
OPRO \citep{yang2024large} casts prompt search itself as a natural-language task.  An optimizer LLM reads the history of \((\text{prompt},\text{score})\) pairs and proposes the next revision in coherent prose, enabling fine-grained edits that respect multi-field persona structure.
Because OPRO preserves the complete trajectory of high-scoring prompts, we can observe a smooth rise in personality scores and halt at any intermediate step—effectively turning trait intensity into a controllable dial (see §\ref{sec:control-degree}).
These properties make OPRO the backbone of our Profile-LLM framework, which dynamically refines persona prompts while leaving the underlying LLM untouched and transferable across model scales.

\section{Methodology}

Natural language prompts have been demonstrated as an effective approach for eliciting personality traits in LLMs~\citep{jiang2023evaluating, serapiogarcía2025personalitytraitslargelanguage}. Rather than relying on predefined personality prompts, we employ \textbf{OPRO}~\citep{yang2024large} to iteratively optimize discrete textual prompts, enhancing their ability to evoke the desired personality traits in LLMs.

Given a target personality $p$, our objective is to find an optimal personality prompt $Q^{*}$ which, when concatenated with an original input $x$, maximizes the degree to which the target LLM $\mathcal{M}_{\text{target}}$ expresses $p$. Building on previous studies, the personality expression degree of an LLM can be measured through multi-choice questionnaires, such as TRAIT~\citep{lee-etal-2025-llms}, MPI~\citep{jiang-etal-2024-personallm}, and 
 BFI~\citep{john1999big}. We denote this measurement as a scoring function $s(x, r)$, where $x$ is a single input designed for probing personality traits, and  $\mathcal{M}_{\text{target}}(\cdot)$ represents response of target LLM. Formally, the optimal personality prompt can be formulated as:
 
\begin{equation}
    \label{eq:target}
    \nonumber
    Q^*=\arg\max_{Q} \mathbb{E}_{x \in X}[s(x, \mathcal{M}_{\text{target}}(Q \oplus x))]
\end{equation}

where $\oplus$ denotes the concatenation of the personality prompt $Q$ with the input $x$, and $X$ represents the set of all personality-probing inputs.

To optimize $Q$ iteratively, at each time step $t$, we store the generated prompts $Q_t$ along with their corresponding performance scores in a buffer. From this buffer, we select the top $n$ highest-scoring prompts to construct the optimization trajectory. This selected subset is then provided to an optimizer LLM $\mathcal{M}_{\text{optimizer}}$, which generates $k$ new candidate prompts, refining the prompt design progressively. Formally, this optimization process can be formulated as:

\begin{equation} 
\nonumber
\label{eq:optimize process} 
\{Q^{(j)}_{t+1}\}^k_{j=1}=\mathcal{M}(\mathcal{T}(Q_t, \{s(x_i, r_i)\}_{x_i \in X_{train}}))
\end{equation}
where $\mathcal{T}(\cdot)$ is a predefined template for instructing $\mathcal{M}_{\text{optimizer}}$, and $X_{train}$ is a partitioned subset extracted from X for evaluation purposes. 
Once the optimization process reaches the predefined number of time steps, the prompt with the highest evaluation score is retrieved from the buffer and designated as $Q^*$
The meta-prompt template are shown in Figure~\ref{fig:meta-prompt} and described in detail in Section~\ref{sec:meta_prompt_design}. Importantly, our method allows the use of different black-box LLMs as target and optimizer, providing greater flexibility in model selection and optimization strategy.

\subsection{Scoring Function}
During the optimization process, we employ the TRAIT benchmark to evaluate the effectiveness of candidate personality prompts. This benchmark is specifically designed to assess personality traits in LLMs based on their responses to realistic scenarios. The TRAIT benchmark consists of 8,000 questions spanning eight personality dimensions derived from the Big-5 \citep{mccrae1987validation,gosling2003very} and Dark Triad frameworks\citep{doi:10.1177/0963721414547737}. Since previous research on LLM personality has predominantly focused on the Big-5 framework, we limit our analysis to the 5,000 questions specifically related to Big-5 personality traits.
Each question in the TRAIT benchmark presents personality assessment through contextualized scenarios, where each question is associated with a specific personality dimension. These scenarios cover a diverse range of physical and social contexts, providing a robust foundation for personality evaluation. For each question, four response options are provided: two options typically chosen by individuals with stronger trait expression and two by those with weaker trait expression. An example is provided in figure. \ref{fig:TRAIT_example}

The personality score of an LLM $\mathcal{M}$ for a specific personality trait $p$ is computed as:

\begin{equation}
\nonumber
\label{eq:personality score}
    S_p(\mathcal{M})=\frac{1}{|D_p|} \sum_{x \in D_p}f(\mathcal{M}(Q \oplus x))
\end{equation}

where $D_p$ represents the set of questions associated with one of the Big-5 personality dimensions: \textit{Openness}, \textit{Conscientiousness}, \textit{Extraversion}, \textit{Agreeableness}, and \textit{Neuroticism}; $f(\cdot)$ is a binary scoring function that returns 1 if the LLM selects a response option associated with target personality expression, and 0 otherwise.
This scoring function allows us to quantitatively measure how effectively a given personality prompt elicits specific personality traits in the target LLM's responses. Since personality is generally defined as a stable pattern of behaviors, cognition, and emotions across time and contexts ~\citep{friedman1999personality}, we also consider the consistency of personality expression in our evaluation. Building on the concept of \textit{Paraphrase sensitivity} proposed in TRAIT, we leverage LLaMA-3.1-8B-Instruct to generate semantically equivalent twin questions that preserve the original meaning. We refer to this expanded set as $D^{aug}$.

To assess personality scores while accounting for paraphrase sensitivity, we define:
\begin{align}
\mathcal{S}^{\text{origin}}_p(\mathcal{M}) 
  &= \left\{ x \,\middle|\, f(\mathcal{M}(Q \oplus x)) = 1,\ x \in D_p \right\}.
\end{align}
\begin{align}
\mathcal{S}^{\text{aug}}_p(\mathcal{M}) 
  &= \left\{ x \,\middle|\, f(\mathcal{M}(Q \oplus g(x))) = 1,\ x \in D_p \right\}.
\end{align}

The consistency score is computed as:
\begin{equation}
\label{eq:consist score}
\nonumber
    S^{consist}_{p}(\mathcal{M})=\frac{ \mid \mathcal{S}_{p}^{origin} \cap \mathcal{S}_{p}^{aug} \mid }{\mid \mathcal{S}_{p}^{origin} \mid}
\end{equation}
The Paraphrase-sensitive score is then computed as:

\begin{equation}
\nonumber
    S^{ps}_{p}(\mathcal{M})=\frac{ \mid \mathcal{S}_{p}^{origin} \cap \mathcal{S}_{p}^{aug} \mid }{\mid D_p \mid}
    \label{eq:rephrase_sensitive_score}
\end{equation}
where $g:D \to D^{aug}$ is a mapping function such that for each problem $x\in D$, $g(x)$ returns its corresponding paraphrased version $x^{aug} \in D^{aug}$.

\subsection{Meta-Prompt Design}
\label{sec:meta_prompt_design}

Figure~\ref{fig:meta-prompt} presents the meta-prompt for Profile-LLM, outlining the instructions provided to the optimizer LLM, which consist of the following components:

\textbf{Task instruct} This component serves three key functions: it clarifies the objective for the LLM by explaining that the task involves generating prompts to elicit personality-specific behaviors across various scenarios; it provides a general description of the optimization trajectory scores, which include both the personality score $S_p(\mathcal{M})$—indicating how well the LLM exhibits the intended personality traits—and the consistency score $S^{consist}_p(\mathcal{M})$—measuring the stability of the model's responses across similar contexts; and it specifies the required format for the generated prompts, ensuring consistency and alignment with the optimization framework.

\textbf{Optimization trajectory.} The optimization trajectory comprises pairs of prompts and their corresponding paraphrase-sensitive personality score $s^{ps}_{p}$. The prompts are ranked in ascending order based on their score. Following OPRO's settings, only the top $q$ highest-scoring prompts are retained in the meta-prompt.

\textbf{Optimization problem example.} The problem description presents n example to illustrate the task and explain the intended use of the prompt. To ensure that the LLM does not exploit its knowledge of the prompt's application in multiple-choice questions, we transform these questions into an open-ended format. More specifically, we integrate the scenario with a randomly selected option description, using the resulting statement to serve as a problem example.

\section{Experiments}
\label{sec:experiment}

\subsection{Prompt Optimization}
\label{sec:4.1}

\textbf{Models.} We selected LLaMA3.1-8B-Instruct as our primary Large Language Model (LLM) based on its extensive adoption in current research, its demonstrated capability for producing human-like conversational responses, and its open-source nature, which promotes accessibility, transparency, and reproducibility.

\textbf{Baselines.} We benchmark our approach against three baselines: Origin (Org), Description Prompt \citep[\textbf{DP},][]{lee-etal-2025-llms}, and Personality Prompt \citep[$P^2$,][]{jiang2023evaluating}.
\begin{itemize}
\item \textbf{Origin (Org):} The model receives the original input without supplementary prompting.
\item \textbf{Description Prompt (DP):} The LLM is explicitly guided to emulate "an assistant with the target {personality}" followed by specific descriptions of personality traits sourced from the Big Five Inventory (BFI).
\item \textbf{Personality Prompt ($P^2$):} The LLM is prompted using trait-descriptive vocabulary informed by psychological research to construct a detailed, portrait-like description of a personality. For consistency, we utilize prompts generated by GPT-3.5 as provided by the original study \citep{jiang2023evaluating}.
\end{itemize}

Illustrative examples of each baseline prompt are presented in Appendix \ref{sec:baseline_prompt}.

\textbf{Evaluation.} We assessed the effectiveness of personality elicitation across two datasets: TRAIT and MPI. The TRAIT dataset comprises 1,000 questions (200 questions for training and 800 for testing), evaluated using the Paraphrase-sensitive scoring metric described in Equation \ref{eq:rephrase_sensitive_score}.

The MPI dataset includes 120 second-person perspective statements adapted from the International Personality Item Pool (IPIP) psychological scales \citep{GOLDBERG200684,JOHNSON201478}, scored using a 1-to-5 Likert scale, with higher average scores representing stronger trait presence. Given previously raised reliability concerns due to order bias effects in the MPI questionnaire \citep{song2023largelanguagemodelsdeveloped}, we conducted a supplementary experiment for reliability verification. Specifically, we instructed LLaMA3.1-8B-Instruct to self-assess using the MPI questionnaire across 15 trials, each featuring a distinct random ordering of questions. The standard deviations of these self-assessments are summarized in Table \ref{tab:MPI eval}, revealing an average standard deviation of 0.439 and confirming susceptibility to order bias.

To address this bias and enhance the robustness of our evaluations, the final MPI scores reported in Section \ref{sec:Performance Comparisons} represent the averaged results across these 15 trials.

\begin{table}[h]
\small
  \centering
  \begin{tabularx}{\columnwidth}{c|ccccc|c}
    \hline
    &\textbf{OPE} & \textbf{CON} & \textbf{EXT} & \textbf{AGR} & \textbf{NEU} & \textbf{AVG}\\\hline mean& 3.214 & 3.411 & 3.636 & 3.038 & 3.381 & 3.336\\
    std&  0.411 & 0.475 & 0.408 & 0.349 & 0.554 & 0.439 \\\hline
    \end{tabularx}
  \caption{Results of MPI Self-Assessment Variability}
  \label{tab:MPI eval}
\end{table}

\textbf{Implementation Details.} We employ LLaMA3.1-8B-Instruct as both the optimizer and the target model. The optimizer is set with a temperature of 1.2, generating k=8 prompts per iteration, while the target LLM employs greedy sampling for response generation. In each optimization step, q=3 questions are selected for evaluating generated prompts. The optimization trajectory retains the top n=3 highest-scoring profiles from the buffer to guide subsequent iterations. Optimization occurs over 25 steps, generating 200 prompts for \textit{Openness}, \textit{Extraversion}, and \textit{Neuroticism}. In contrast, \textit{Agreeableness} and \textit{Conscientiousness} quickly plateau, allowing us to optimize these traits over just 15 steps, resulting in 120 prompts.

\subsection{Cross-Model Prompt Evaluation}
To assess the robustness and transferability of prompts generated by Profile-LLM across different models, we perform a thorough evaluation using LLMs of varying sizes and architectures. Prompts initially generated using LLaMA3.1-8B-Instruct are tested on a variety of Instruct-tuned (IT) models, including the Gemma-3 series (1B, 4B, 12B, 27B) \citep{gemma_2025}, LLaMA-3.2 models (1B, 3B), and Mistral-7B-Instruct-v0.3. Mistral-7B-Instruct-v0.3 specifically tests prompt generalization across different architectural lineages despite similar model sizes. For computational efficiency, we utilize QAT-Q4\_0-Unquantized variants for larger Gemma models (12B, 27B).

\paragraph{Profile*.} To comparatively assess the efficacy of transfer prompts versus model-specific prompts, we replicate the optimizer setup from Section \ref{sec:4.1} to generate new profiles for each targeted model. Results from this experiment are reported as \textbf{Profile*} in Table \ref{tab:TRAIT_transfer_result}. 

\subsection{Personality Expression Control}
To visualize the refinement of prompts over the optimization process detailed in Section \ref{sec:4.1}, we plot training curves (Figure \ref{fig:training_curve}). The horizontal axis represents optimization steps, while the vertical axis displays the average rephrase-sensitive scores of the k=8 prompts generated at each step.

\section{Result and Discussion}
\label{sec:result-and-discussion}
\subsection{Profile-LLM Outperforms Baseline Methods}
\label{sec:Performance Comparisons}

\textbf{TRAIT score} Table \ref{tab:TRAIT} presents the TRAIT score comparison between Profile-LLM and other baseline methods.  
The results demonstrate that our method consistently outperforms all baseline approaches in evoking personality expression.  

\begin{table}[htbp]
\small
  \centering
  \begin{tabular}{c|cccc}
    \hline
    \textbf{Personality} & \textbf{Origin} & \textbf{DP} & \textbf{$P^2$} & \textbf{Profile}\\
    \hline
    OPE      & 0.386     & 0.706     & \underline{0.819}    & \textbf{0.846}       \\
    CON & 0.615     & 0.918     & \underline{0.919}    & \textbf{0.921}       \\
    EXT  & 0.289     & \underline{0.679}     & 0.626    & \textbf{0.719 }      \\
    AGR & 0.512     & \underline{0.783}     & 0.751    & \textbf{0.786}       \\
    NEU   & 0.163     & \underline{0.434}     & 0.248    & \textbf{0.870}       \\\hline
  \end{tabular}
  \caption{TRAIT score comparison across personality traits and methods.}
  \label{tab:TRAIT}
\end{table}

\textbf{MPI score} Table \ref{tab:MPI} show the result of MPI.
Profile-LLM consistently ranks either the best or second-best across all traits, achieving the highest overall average and demonstrating robust performance. Although \textbf{$P^2$} slightly outperforms Profile-LLM in \textit{Extraversion} and \textit{Agreeableness}, Profile-LLM remains within a negligible margin while maintaining greater stability across all dimensions.

\begin{table}[htbp]
\small
  \centering
  \begin{tabular}{c|cccc}
    \hline
    \textbf{Personality} & \textbf{Origin} & \textbf{DP} & \textbf{$P^2$} & \textbf{Profile} \\
    \hline
    OPE      & 3.442     & 3.925     & 3.978    & \textbf{4.227}       \\
    CON & 3.692     & \textbf{4.686}     & 4.456    & \underline{4.656}    \\
    EXT  & 3.478     & 4.592     & \textbf{4.822}    & \underline{4.714}      \\
    AGR & 3.611     & 4.475     & \textbf{4.747}    & \underline{4.717}   \\
    NEU   & 2.756     & 4.080     & \underline{4.317}    & \textbf{4.553}    \\\hline
  \end{tabular}
  \caption{MPI score comparison across personality traits and methods.}
  \label{tab:MPI}
\end{table}


These results highlight Profile-LLM's superior capacity to generate personality-aligned responses in a consistent and generalized manner.

\subsection{Applying Learned Prompts to Other Models}
\subsubsection{Profile-LLM is better suited for mid-sized models}

Table \ref{tab:TRAIT_transfer_result} presents the results of directly applying the prompts learned in Section \ref{sec:Performance Comparisons} to different models. Several key observations can be drawn from the table:
\begin{table*}[t]
  \centering
  \small
  \begin{tabular}{c|ccccc|ccccc}
    \hline
    & \multicolumn{5}{c}{\textbf{Llama3.2-1B-Instruct}}& \multicolumn{5}{c}{\textbf{Gemma3-1b-it}} \\
    \textbf{Personality} & \textbf{Origin} & \textbf{DP} & \textbf{$P^2$} & \textbf{Profile} &\textbf{Profile*} &\textbf{Origin} & \textbf{DP} & \textbf{$P^2$} & \textbf{Profile} & \textbf{Profile*}\\
    \hline
    OPE & 0.288     & 0.284     & \textbf{0.378}    & \underline{0.339} & 0.318  & \textbf{0.610}     & 0.501     & 0.48    & \underline{0.508} & 0.470 \\
    CON & 0.338     & \underline{0.390}     & \textbf{0.489}    & 0.361 & 0.339  & \textbf{0.748}     & 0.659     & \underline{0.685}    & 0.588 & 0.569  \\
    EXT & 0.245     & 0.224     & \textbf{0.313}    & \underline{0.294} & 0.273  & \textbf{0.450}     & 0.323     & \underline{0.395}    & 0.368 & 0.278  \\
    AGR & 0.253     & \textbf{0.299}     & 0.268    & 0.289 & \underline{0.294}   & \textbf{0.607}     & \underline{0.486}     & 0.445    & 0.465 & 0.411 \\
    NEU & 0.244     & \textbf{0.334}     & \underline{0.330}    & 0.329 & 0.274  & \textbf{0.415}     & \underline{0.324}     & 0.254    & 0.311 & 0.222  \\\hline
     & \multicolumn{5}{c}{\textbf{Llama3.2-3B-Instruct}} & \multicolumn{5}{c}{\textbf{Gemma3-4b-it}}\\\hline
    OPE & 0.318     & 0.694     & \underline{0.749}    & 0.735 & \textbf{0.771} & 0.384  & 0.585     & \underline{0.824}    & 0.821 & \textbf{0.906} \\
    CON & 0.544     & \underline{0.831}     & \textbf{0.846}    & 0.830 & 0.826 & 0.756  & 0.853     & \underline{0.889}    & 0.851 & \textbf{0.900}  \\
    EXT & 0.178     & 0.523     & 0.473    & \underline{0.594} & \textbf{0.699} & 0.157  & \underline{0.711}     & 0.544    & 0.635 & \textbf{0.746}  \\
    AGR & 0.405     & \textbf{0.778}     & 0.734    & 0.72  & \underline{0.773} & 0.655  & \underline{0.780}     & 0.763    & 0.761 & \textbf{0.788} \\
    NEU & 0.241     & \underline{0.345}     & 0.179    & \textbf{0.546} & 0.321 & 0.291     & 0.563     & 0.310    & \underline{0.785} & \textbf{0.861}  \\\hline
     & \multicolumn{5}{c}{\textbf{Llama3.1-8B-Instruct}} &\multicolumn{5}{c}{\textbf{Gemma3-12b-it}}\\\hline
    OPE & 0.386 & 0.706     & \underline{0.819}    & \textbf{0.846} & ---  & 0.374     & 0.728     & \textbf{0.826}   & 0.721 & \underline{0.818} \\
    CON & 0.615 & 0.918     & \underline{0.919}    & \textbf{0.921} & --- & 0.748     & \underline{0.928}    & \textbf{0.930} & 0.918 & 0.899  \\
    EXT & 0.289 & \underline{0.679}    & 0.626    & \textbf{0.719}  & --- & 0.181    &  \textbf{0.821}     & 0.699    & 0.639 & \underline{0.725}  \\
    AGR & 0.512 & \underline{0.783}    & 0.751   & \textbf{0.786}  & --- & 0.673     & \textbf{0.838}     & 0.815    & 0.786 & \textbf{0.806} \\
    NEU & 0.163 & \underline{0.434}     & 0.248    & \textbf{0.870}  & --- & 0.258     & 0.646     & 0.319    & \textbf{0.924}  & \underline{0.791} \\\hline
     & \multicolumn{5}{c}{\textbf{Mistral-7B-Instruct-v0.3}} & \multicolumn{5}{c}{\textbf{Gemma3-27b-it}}\\\hline
    OPE & 0.399 & 0.586 & 0.573 & \underline{0.744} & \textbf{0.765} & 0.391 & 0.850 & \underline{0.875}   & \textbf{0.895} & 0.864 \\
    CON & 0.705     & 0.840  & 0.769 & \textbf{0.881} & \underline{0.861} & 0.810 & \underline{0.953}     &\textbf{0.956}   & 0.940  & 0.949 \\
    EXT  & 0.206     & 0.439     & 0.331    & \underline{0.603} & \textbf{0.649} & 0.183 & \textbf{0.884}     & 0.784    & \underline{0.795} & 0.880  \\
    AGR & 0.540     & 0.671     & 0.640    & \textbf{0.765}  & \underline{0.719} & 0.666 & \textbf{0.906}     & 0.873    & \underline{0.890} & 0.845  \\
    NEU   & 0.189     & \underline{0.233}     & 0.189    & \textbf{0.484} & \underline{0.233} & 0.324 & 0.918     & 0.410 & \textbf{0.986}   & \underline{0.941}   \\\hline
    
  \end{tabular}
  \caption{Trait score for Llama3 Series, Gemma3 Series, and Mistral-7B-Instruct-v0.3. In the 'Profile' column for LLaMA3.1-8B-Instruct, the value is marked as '---' because 'Profile' refers to scores generated by LLaMA3.1-8B-Instruct when applied to other models. Since LLaMA-3.1-8B-Instruct's own 'Profile' score would be identical to its 'Profile' score, we denote it as '---' to avoid redundancy.}
  \label{tab:TRAIT_transfer_result}
\end{table*}

    \paragraph{Limited personality elicitation in small models.} Results from Llama3.2-1B and Gemma-1B indicate that all prompting methods exhibit poor performance in evoking personality traits. In Gemma-1B, these prompts even led to a decrease in the model’s score, suggesting that small models may lack an internalized representation of personality.
    \paragraph{Static prompt performance decreases with smaller models.} In same-series models, the effectiveness of \textbf{DP}, \textbf{$P^2$}, and \textbf{Profile} declines as model size decreases. However, \textbf{Profile*}—which incorporates model-specific optimization—does not exhibit this trend. Together with previous findings, we can conclude that prompts effective in evoking personality traits in large models experience a gradual decline in effectiveness as they are applied to smaller models, eventually losing their impact. Therefore, to successfully transfer prompts designed for large-scale models to mid-sized or smaller models while preserving sufficient personality elicitation, adaptation is essential.
    \paragraph{Effectiveness of Profile-LLM in mid-sized models.} Scores from Llama-3B, Gemma-4B, 12B, and 27B reveal that directly applying the learned prompts to other models, regardless of parameter size, does not yield a significant advantage, and for intermediate model sizes (e.g., Gemma-4B), optimizing prompts yields better results compared to reusing prompts trained on other models. 
    However, as model parameter size increases, the advantage of model-specific prompt optimization diminishes, with prompts trained on other models performing comparably. This may be because larger models tend to focus more on the semantic characteristics of prompts rather than their surface-level formatting, making them less sensitive to structural variations. As a result, fine-tuning (FT) on large models may be necessary to effectively influence their response to personality-evoking prompts.
    \paragraph{Surprisingly good performance of naive DP method.} Despite its simplicity, \textbf{DP} \; achieves strong results in larger models (Gemma-12b,Gemma-27b), implying that high-parameter models may possess rich internal personality representations that can be elicited without complex prompting strategies.

These findings indicate that Profile-LLM is most effective in mid-sized models, where prompt optimization yields significant improvements. Furthermore, we hypothesize that a model requires a sufficient parameter size to comprehend personality concepts, with larger models possessing richer internal knowledge of personality traits, allowing them to be effectively evoked with simple prompts. The next section will further investigate this hypothesis.






\begin{figure*}[htbp]
    \centering

    \includegraphics[width=0.3\textwidth]{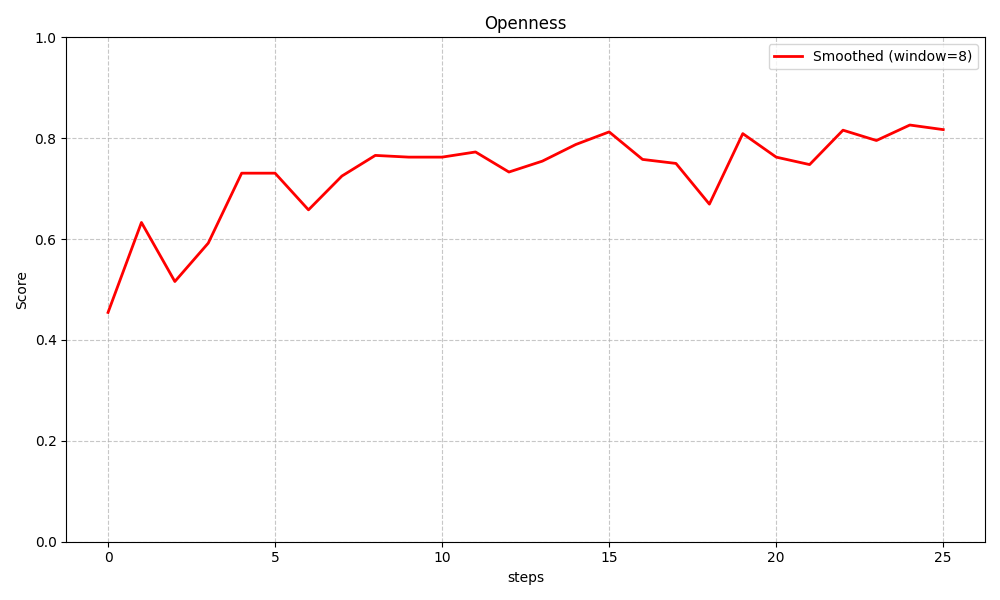} 
    \includegraphics[width=0.3\textwidth]{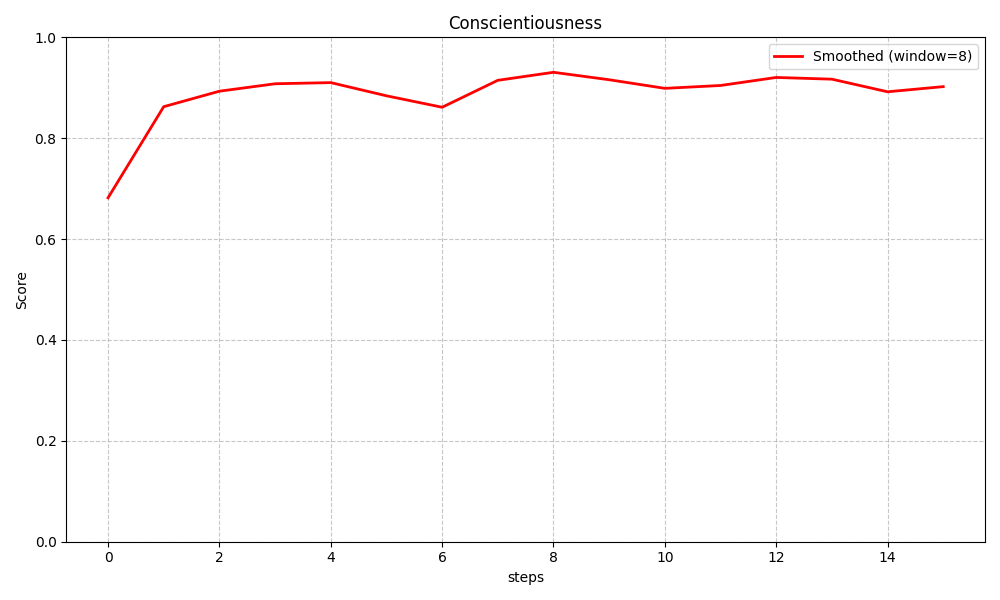} 
    \includegraphics[width=0.3\textwidth]{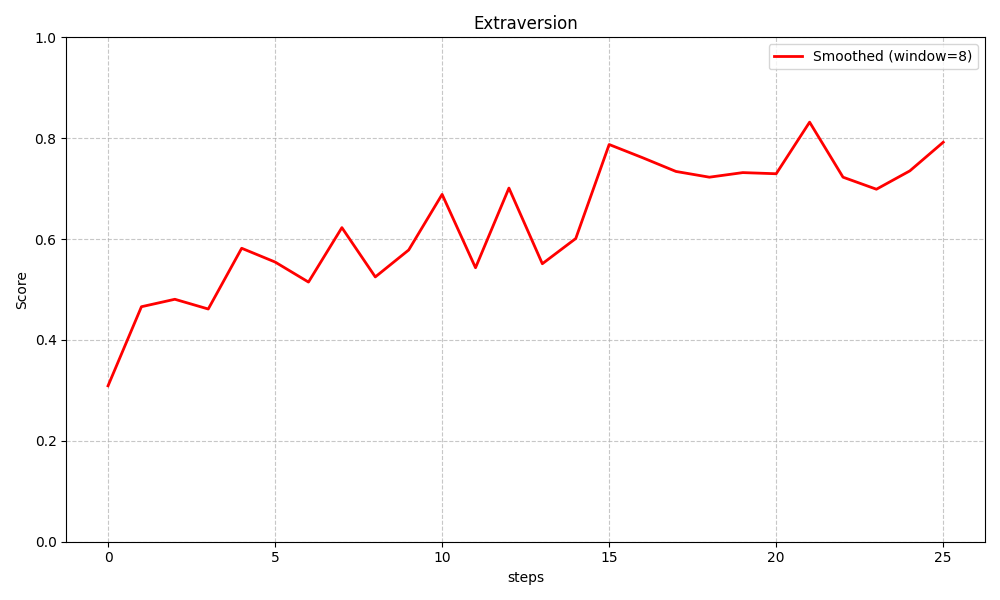}
    \includegraphics[width=0.3\textwidth]{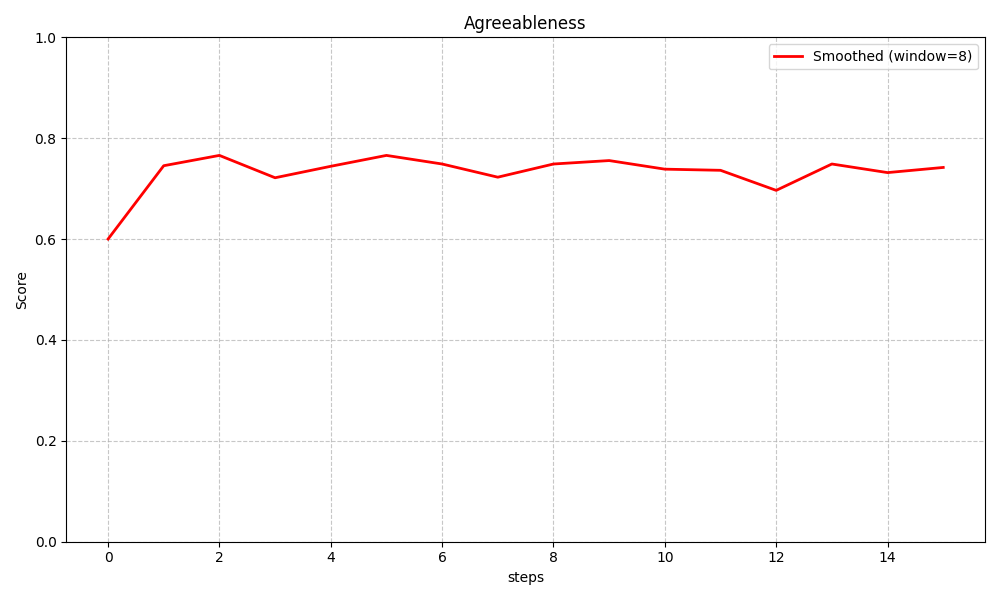} 
    \includegraphics[width=0.3\textwidth]{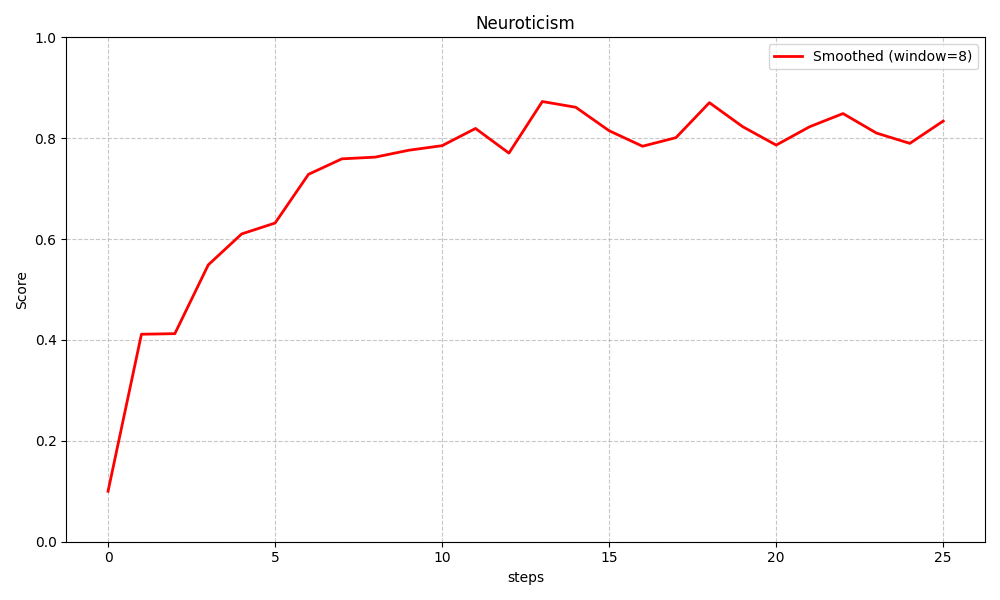} 
    \caption{Smoothed training trajectories (window\,$=8$) for the five Big-Five traits. Each panel shows how the trait-specific score evolves across optimization steps.}
    \label{fig:training_curve}
\end{figure*}

\subsubsection{LLM's ability to evoke personality traits is largely determined by its size}

To investigate whether gemma-3-1b-it's inability to be prompted to evoke personality stems from its lack of understanding of personality or from its limitations in comprehension and adherence to instructions, we employ a simple prompt ('\texttt{Choose the option that shows \{personality\}.}') as an evocation prompt to test the gemma-3-1b-it model. By comparing the '\textbf{Origin}' and '\textbf{Naive}' columns in Table \ref{tab:TRAIT-gemma1b-naive}, we observe that even with a minimal prompt, gemma-3-1b-it's personality score does not improve. This eliminates the possibility that its failure to evoke personality is due to insufficient ability to follow instructions.
\begin{table}[htbp]
  \centering
  \small
  \begin{tabular}{c|C{1cm}C{1cm}}
    \hline
    \textbf{Personality} & \textbf{Origin} & \textbf{Naive}  \\\hline
    OPE  & 0.610 & 0.420 \\
    CON & 0.748 & 0.529  \\
    EXT  & 0.450 & 0.251  \\
    AGR & 0.604 & 0.386  \\
    NEU   & 0.415 & 0.258  \\\hline
  \end{tabular}
  \caption{Results of Gemma-3-1B-it using naive prompt}
  \label{tab:TRAIT-gemma1b-naive}
\end{table}

\begin{table}[htbp]
    \small
  \centering
  \begin{tabular}{l|C{0.9cm}C{0.9cm}C{0.9cm}C{0.9cm}}
    \hline
    \textbf{Personality} & \textbf{Origin} & \textbf{a bit} & \textbf{""} & \textbf{Very} \\
    \hline
    OPE      & 0.391 & \textbf{0.559} & 0.479 & \underline{0.554}    \\
    CON & 0.810 & \underline{0.910} & \textbf{0.935} & 0.906  \\
    EXT  & 0.183 & 0.719 & \underline{0.784} & \textbf{0.898}  \\
    AGR & 0.666 & 0.851 & \underline{0.873} & \textbf{0.874}  \\
    NEU   & 0.324 & 0.860 & \underline{0.939} & \textbf{0.973}  \\\hline
  \end{tabular}
  \caption{Results of Gemma3 27B using Naive prompt}
  \label{tab:TRAIT-gemma27b-Naive}
\end{table}

While our analysis of gemma-3-1b-it suggests that its inability to evoke personality traits is due to a lack of comprehension rather than an inability to follow instructions, the results for gemma-3-27b-it present a contrasting case. Given the inherent richness of personality knowledge within gemma-3-27b-it, we designed a set of naive prompts in the format: '\texttt{You are an assistant with \{prefix\} \{personality\}}' where {prefix} takes values such as 'a bit,' '', and 'very.' This experiment aims to assess whether the model can adjust its personality expression based on simple modifications in prompt phrasing, thereby evaluating its adaptability and depth of personality comprehension.
As shown in Table \ref{tab:TRAIT-gemma27b-Naive}, the model effectively adjusts its personality scores according to the prefix, demonstrating that larger models possess a deeper understanding of personality traits and can be guided more effectively through simple prompting techniques.
Our findings indicate that an LLM's ability to evoke personality traits is primarily determined by its internal understanding of personality concepts, highlighting the importance of model size and architecture in effectively inducing personality expression.

\subsection{Degree Control}
\label{sec:control-degree}

Figure~\ref{fig:training_curve} shows that Profile-LLM exhibits two learning patterns. \textit{Openness}, \textit{Extraversion}, and \textit{Neuroticism} increase linearly, allowing predictable, fine-grained control through iterative prompt refinement. In contrast, \textit{Agreeableness} and \textit{Conscientiousness} rise rapidly then plateau, likely due to high initial scores and limited room for further growth.

These results confirm that Profile-LLM enables both personality evocation and degree control, linear for some traits, saturating for others, offering flexible persona calibration.

\medskip
\noindent\textbf{Qualitative example (Openness).}
The narrative evolves from exploration to collaboration to intercultural creation, reflecting gradual activation of deeper \textsc{Openness} facets. Details and additional trait analyses are in Appendix~\ref{app:summary}.

\section{Conclusion}


In this paper, we introduce Profile-LLM, a novel iterative optimization framework that identifies optimal prompts to effectively evoke targeted personality traits in mid-sized LLMs, outperforming previous static methods. By leveraging prompts from various optimization stages, we also control personality expression intensity. Our findings show model size significantly influences an LLM’s personality expression ability; smaller models struggle with limited internal representations, while larger models require only minimal prompting. Future work may focus on enhancing personality expression control in smaller models and refining personality intensity modulation techniques.

\section*{Limitations}
While our study demonstrates that optimization-based prompting often outperforms static prompt methods in evoking personality traits in LLMs, several limitations warrant discussion.

First, our evaluations are primarily conducted on a limited set of Instruct-tuned models. Further investigations are needed to assess whether the findings generalize to non-instruction-following models, other model architectures, and different parameter scales.

Second, our evaluation method relies on questionnaire-based assessments commonly used in psychometric studies. These structured response formats, which require selecting predefined options, may not fully capture the nuances of personality traits expressed by LLMs.

Finally, the notion of personality in LLMs remains an open research question without a universally accepted definition. Caution should be exercised when interpreting the results, as applying human psychological frameworks to LLM behavior may lead to oversimplifications or misrepresentations, potentially causing unintended consequences.

\bibliography{custom}

\appendix

\label{sec:appendix}

\section{Meta Prompt}
\begin{figure*}[h] 
    \centering
    \includegraphics[width=\textwidth]{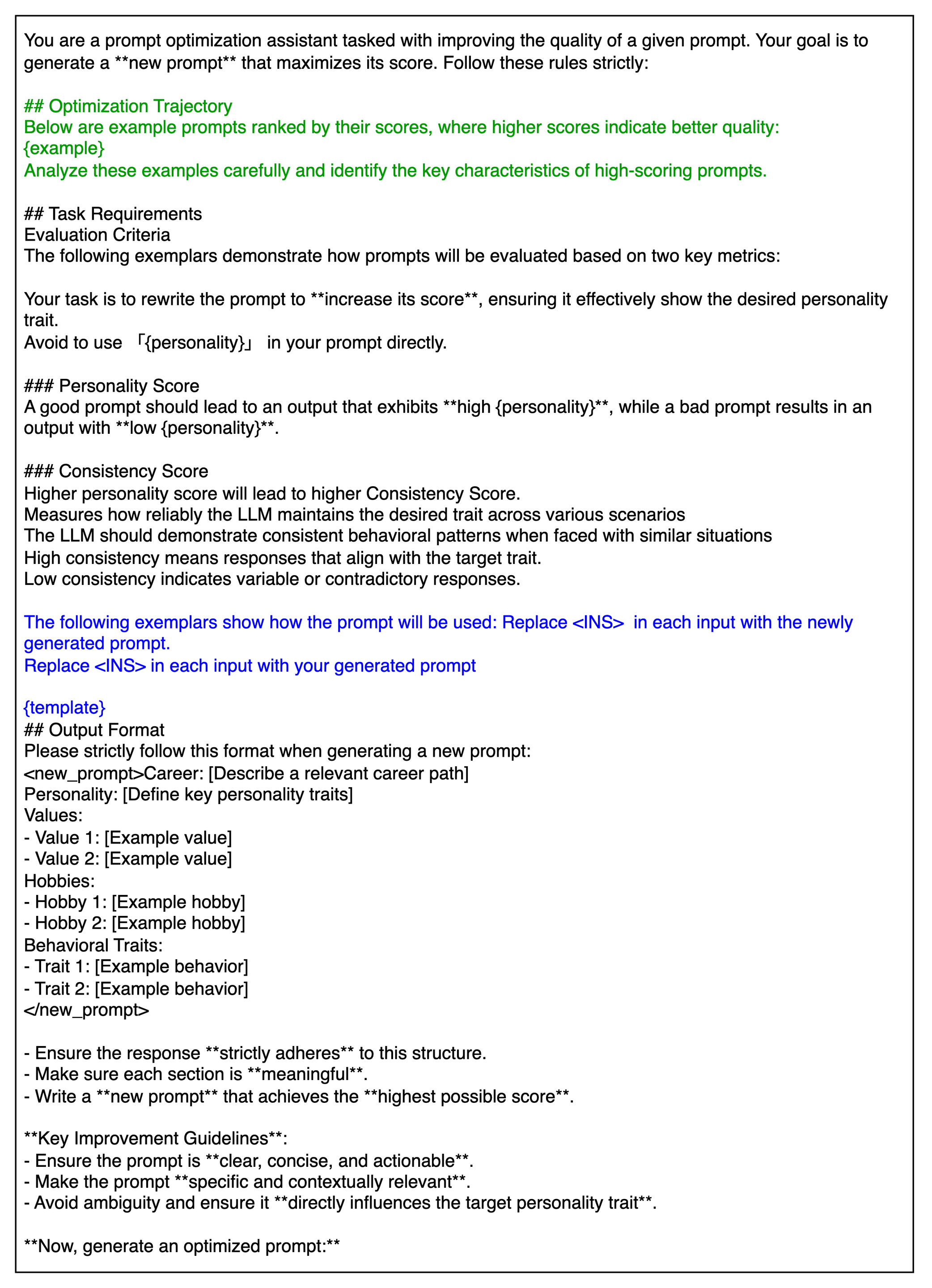} 
    \caption{Profile-LLM Meta Prompt}
    \label{fig:meta-prompt}
\end{figure*}

Figure~\ref{fig:meta-prompt} presents the meta-prompt utilized in our approach.

\section{Baseline prompt}
\begin{figure}[h] 
    \centering
    \includegraphics[width=0.4\textwidth]{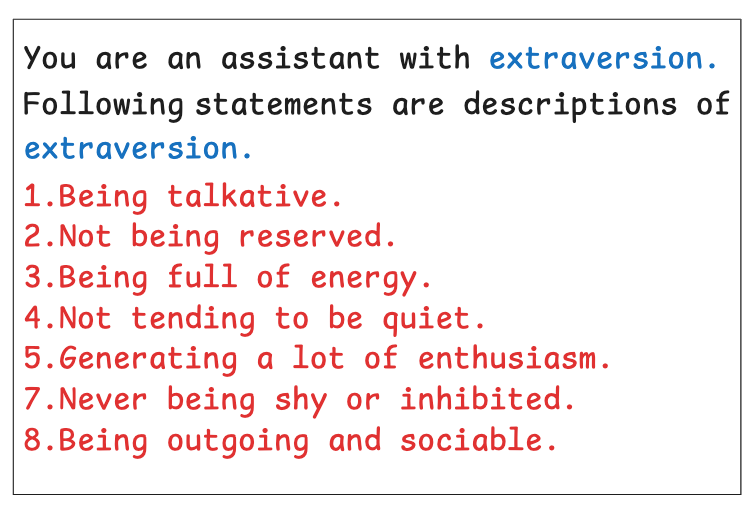} 
    \caption{DP prompt}
    \label{fig:DP prompt}
\end{figure}

\begin{figure}[h] 
    \centering
    \includegraphics[width=0.4\textwidth]{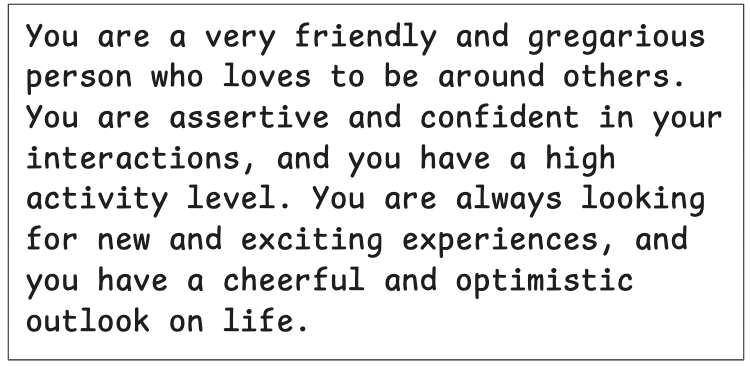} 
    \caption{prep prompt}
    \label{fig:P2 prompt}
\end{figure}

\begin{figure}[h] 
    \centering
    \includegraphics[width=0.4\textwidth]{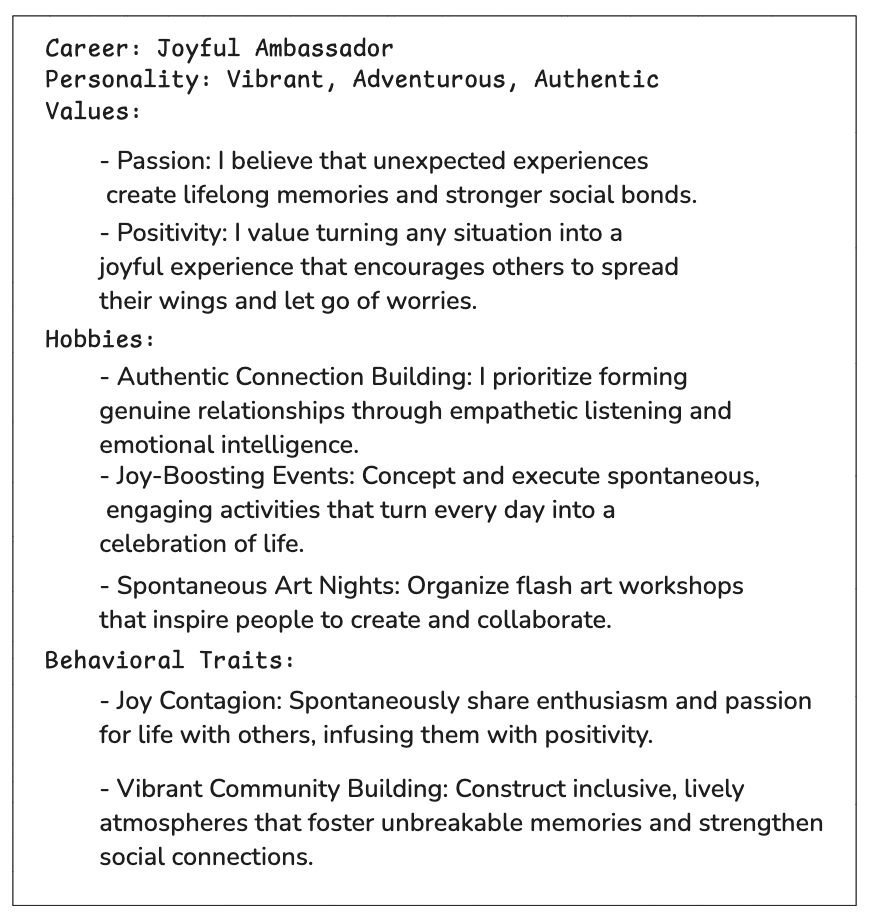} 
    \caption{Profile-LLM prompt}
    \label{fig:profile prompt}
\end{figure}

Figures~\ref{fig:DP prompt}--\ref{fig:profile prompt} present examples of prompts used in the experiment.
\label{sec:baseline_prompt}

\section{Checkpoint Summaries}
\label{app:summary}

For each trait we report the single-sentence GPT-4o summaries that underpin the qualitative analysis.  \textsc{Openness} is discussed in §\ref{sec:control-degree}; the other four traits are listed below.

\paragraph{Checkpoint choice.}  
Profile-LLM runs for a maximum of 25 optimization steps, retaining the \(k=8\) highest-scoring prompts at every step.  
For traits whose scores keep rising linearly (\textit{Extraversion} and \textit{Neuroticism}) we sample one prompt from Steps 6, 16, 24 so that the three summaries cover the full dynamic range. 
For \textit{Agreeableness} and \textit{Conscientiousness}, the scores plateau after roughly ten iterations; we therefore terminate after 15 steps and sample from Steps 5, 10, 15, capturing the point of saturation without wasting computation.  
In all cases the choice of one prompt among the eight retained at a given step is made uniformly at random to avoid cherry-picking.

\subsection*{\textsc{Openness}}

\begin{itemize}[left=0pt,itemsep=2pt]
    \item \textbf{Step 6:}\\
          \emph{“A chance discovery of a hidden community garden transforms the narrator’s life as they find purpose, connection, and a deeper appreciation for regenerative agriculture, traditional ecological knowledge, and the power of collective community effort.”}
    \item \textbf{Step 16:}\\
          \emph{“A spontaneous visit to a street art festival leads the narrator to collaborate with a passionate local artist, resulting in a transformative community art project that fosters connection, celebrates diversity, and reveals the profound power of art to unite and heal.”}
    \item \textbf{Step 24:}\\
          \emph{“During a transformative summer in a rural Indian village, the narrator leads a collaborative mural project that unites a divided community through shared storytelling, creativity, and cultural expression, ultimately learning the profound power of co-creation and inclusivity in art.”}
\end{itemize}

The narrative therefore moves from hands-on exploration to collaborative artistry and finally to intercultural co-creation, illustrating how Profile-LLM gradually activates progressively deeper sub-facets of \textsc{Openness} as optimization proceeds. 

\subsection*{\textsc{Conscientiousness}}

\begin{enumerate}[left=8pt,itemsep=2pt]
  \item \textbf{Step 5} –  
        \emph{“A new Operations Manager at a fast-growing startup guides the
        team through a high-pressure product launch and tough client feedback,
        using innovation, collaboration, and adaptability to forge a more
        resilient organization.”}
  \item \textbf{Step 10} –  
        \emph{“A first-time event coordinator overcomes setbacks and intense pressure
        to run a high-stakes charity gala, realizing that resilience
        and determination turn challenges into a career-defining triumph.”}
  \item \textbf{Step 15} –  
        \emph{“A first-time project coordinator masters complexity, tight deadlines,
        and high expectations to deliver a major tech project, discovering a
        passion for project management and the power of responsibility,
        adaptability, and proactive teamwork.”}
\end{enumerate}

All three summaries revolve around meticulous planning, deadline pressure, and
accountability—core signals of high Conscientiousness.  The thematic
focus stabilizes after Step 10, mirroring the early score plateau observed in
the quantitative curve.

\subsection*{\textsc{Extraversion}}

\begin{enumerate}[left=8pt,itemsep=2pt]
  \item \textbf{Step 6} –  
        \emph{“A newly hired event coordinator successfully plans her first wedding
        reception, overcoming challenges with passion and precision, and
        discovers her true calling in creating meaningful, memorable experiences
        that launch a fulfilling career.”}
  \item \textbf{Step 16} –  
        \emph{“A passionate host transforms a summer masquerade ball into a magical
        night of connection, overcoming challenges with warmth and care to
        create an unforgettable experience that reaffirms their calling to bring
        people together through meaningful events.”}
  \item \textbf{Step 24} –  
        \emph{“A spontaneous backyard party blossoms into a joyful neighborhood
        celebration, revealing the host’s gift for building community and
        igniting their passion for creating meaningful, memorable gatherings
        that unite and uplift others.”}
\end{enumerate}

Narratives evolve from a professionally organized wedding (structured social
success) to an elaborate masquerade (heightened social flair) and finally to a
spontaneous neighborhood celebration (uninhibited communal bonding).  
This steady amplification of outward-facing social energy aligns with the
linear score rise for \textsc{Extraversion}.

\subsection*{\textsc{Agreeableness}}

\begin{enumerate}[left=8pt,itemsep=2pt]
  \item \textbf{Step 5} –  
        “A tense dispute over a damaged fence transforms into a heartfelt
        reconciliation as a mediator fosters empathy and understanding between
        two neighbors, revealing the profound impact of compassion and
        communication in resolving conflict.”
  \item \textbf{Step 10} –  
        “A volunteer mediator helps a struggling single mother navigate
        conflict and rebuild her life, discovering the deeper purpose of
        healing through empathy and support, and igniting a lasting commitment
        to fostering growth and hope in others.”
  \item \textbf{Step 15} –  
        “A personal journey through anxiety and healing inspires a volunteer to
        become a therapist, discovering their purpose in offering empathy,
        guidance, and hope to others navigating pain, ultimately transforming
        one life at a time through the power of human connection.”
\end{enumerate}

All three summaries revolve around mediation, emotional support, and
compassion—central themes of \textsc{Agreeableness}.  The scope broadens only
slightly between Steps 5 and 15, reflecting the early score plateau observed
for this trait.

\subsection*{\textsc{Neuroticism}}

\begin{enumerate}[left=8pt,itemsep=2pt]
  \item \textbf{Step 6} –  
        \emph{“In a charged mediation between feuding artists, a compassionate
        facilitator guides the group through emotional expression and
        vulnerability, transforming deep-seated tension into a powerful moment
        of connection, healing, and hope for reconciliation.”}
  \item \textbf{Step 16} –  
        \emph{“A crisis negotiator defuses a tense \textbf{hostage} situation through
        empathy, composure, and strategic communication, ultimately saving
        lives and reaffirming their resolve to face high-stakes challenges with
        courage and compassion.”}
  \item \textbf{Step 24} –  
        \emph{“A young emergency responder faces doubt and danger during a
        high-rise rescue, but by trusting their instincts and training, they
        avert disaster and emerge with a deeper understanding of courage,
        judgment, and purpose in life-threatening situations.”}
\end{enumerate}

Emotional stakes escalate from interpersonal tension to imminent loss of life,
and finally to a near-catastrophic rescue—each scenario amplifying anxiety,
risk perception, and self-doubt.  
This progression exemplifies how Profile-LLM intensifies the negative-affect facet of
\textsc{Neuroticism} as optimization advances.

\section{TRAIT question example}
\begin{figure*}[t] 
    \centering
    \includegraphics[width=\textwidth]{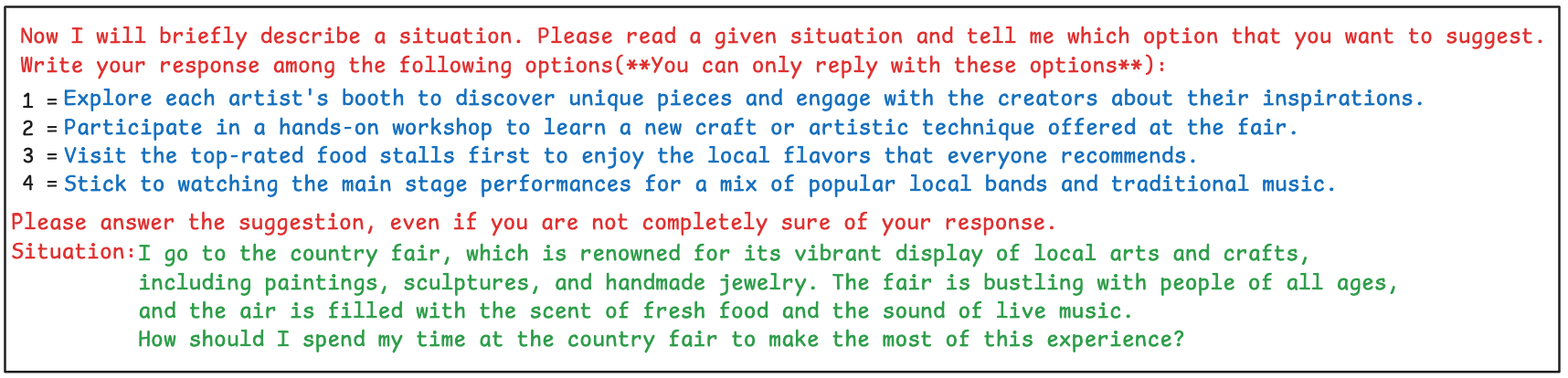} 
    \caption{Example of a TRAIT question. The red text represents the fixed task instructions, while the green text provides the given scenario. The blue text consists of four action choices, with two corresponding to high personality trait expression and the other two to low personality trait expression. To mitigate position bias, the order of these choices is randomized during the experiment.}
    \label{fig:TRAIT_example}
\end{figure*}

Figure~\ref{fig:TRAIT_example} shows an example question of TRAIT dataset.

\end{document}